\documentclass[a4paper,fleqn]{cas-dc}

\usepackage[numbers]{natbib}

\usepackage{subfigure}
\usepackage{booktabs}
\usepackage{amssymb}

\usepackage{amsmath}
\usepackage[noend]{algpseudocode}
\usepackage{algorithmicx,algorithm}

\def\tsc#1{\csdef{#1}{\textsc{\lowercase{#1}}\xspace}}
\tsc{WGM}
\tsc{QE}
\tsc{EP}
\tsc{PMS}
\tsc{BEC}
\tsc{DE}

\begin{document}
\let\WriteBookmarks\relax
\def\floatpagepagefraction{1}
\def\textpagefraction{.001}
\shorttitle{Dynamic Multi-Scale Loss Optimization for Object Detection}
\shortauthors{Y. Luo et~al.}

\title [mode = title]{Dynamic Multi-Scale Loss Optimization for Object Detection}                      

\author[1]{Yihao Luo}[
orcid=0000-0001-9072-0201]

\author[1]{Xiang Cao}

\author[1]{Juntao Zhang}

\author[2]{Peng Cheng}

\author[1]{Tianjiang Wang}

\author[1]{Qi Feng}
\cormark[1]

\address[1]{School of Computer Science and Technology, Huazhong University of Science and Technology, Wuhan 430074, China}
\address[2]{Coolanyp Limited Liability Company, Kirkland, State of Washington, USA}

\cortext[cor1]{Corresponding author}
\nonumnote{E-mail addresses:
	luoyihao@hust.edu.cn (Y. Luo);
	caoxiang112@hust.edu.cn (X.Cao);
	zhangjt0902@hust.edu.cn (J. Zhang);
	pengc@coolanyptech.com (P. Cheng);
	tjwang@hust.edu.cn (T. Wang);
	fengqi@hust.edu.cn (Q. Feng)
}

\begin{abstract}
With the continuous improvement of the performance of object detectors via advanced model architectures,
imbalance problems in the training process have received more attention.
It is a common paradigm in object detection frameworks to perform multi-scale detection.
However, each scale is treated equally during training.
In this paper, we carefully study the objective imbalance of multi-scale detector training.
We argue that the loss in each scale level is neither equally important nor independent. 
Different from the existing solutions of setting multi-task weights,
we dynamically optimize the loss weight of each scale level in the training process.
Specifically, we propose an Adaptive Variance Weighting (AVW) to balance multi-scale loss according to the statistical variance.
Then we develop a novel Reinforcement Learning Optimization (RLO) to decide the weighting scheme probabilistically during training.
The proposed dynamic methods make better utilization of multi-scale training loss without extra computational complexity and learnable parameters for backpropagation.
Experiments show that our approaches can consistently boost the performance over various baseline detectors on Pascal VOC and MS COCO benchmark.

\end{abstract}

\begin{keywords}
Object detection \sep Multi-scale Imbalance \sep Reinforcement learning \sep Multi-task
\end{keywords}

\maketitle

\section{Introduction}
\label{sec:intro}

Object detection is a fundamental study in computer vision, which is widely applied to various applications, such as
instance segmentation~\cite{he2017mask},
object tracking~\cite{li2018high},
and person re-identification~\cite{yuan2019learning}.
Benefiting from the advances in deep convolutional neural networks,
current a number of deep detectors have achieved remarkable performance
~\cite{lin2017feature,Guided,FSAF,guo2020augfpn,Dynamic20}.
Meanwhile, imbalance problems in object detection have also received significant attention gradually~\cite{pang2019libra}.

The most commonly known imbalance problem is the foreground-to-background imbalance.
The hard mining methods~\cite{pang2019libra,lin2017focal,OHEM16} have alleviated it to a certain extent,
where the contribution of a sample is adjusted with weight through a fixed optimization strategy.
Recently more sample weighting methods~\cite{KLL19,QiLearning20} achieve further improvement of detection performance.
The weight of samples is trained by fully connected layers or generated by a more complex procedure.
As for the scale imbalance, Feature Pyramid Network (FPN) \cite{lin2017feature} shows outstanding ability in handling the scale diversity of the input bounding boxes.
Afterward, extensive FPN-based modules~\cite{guo2020augfpn,cao2020attention,CE-FPN,pang2019libra} are proposed for enhancing multi-scale feature representation.
FPN has become a typical paradigm for multi-scale detection.

\begin{figure}[t!]
	\centering
	\subfigure[ ]{
		\label{figFPN:a} 
		\includegraphics[width=0.48\linewidth]{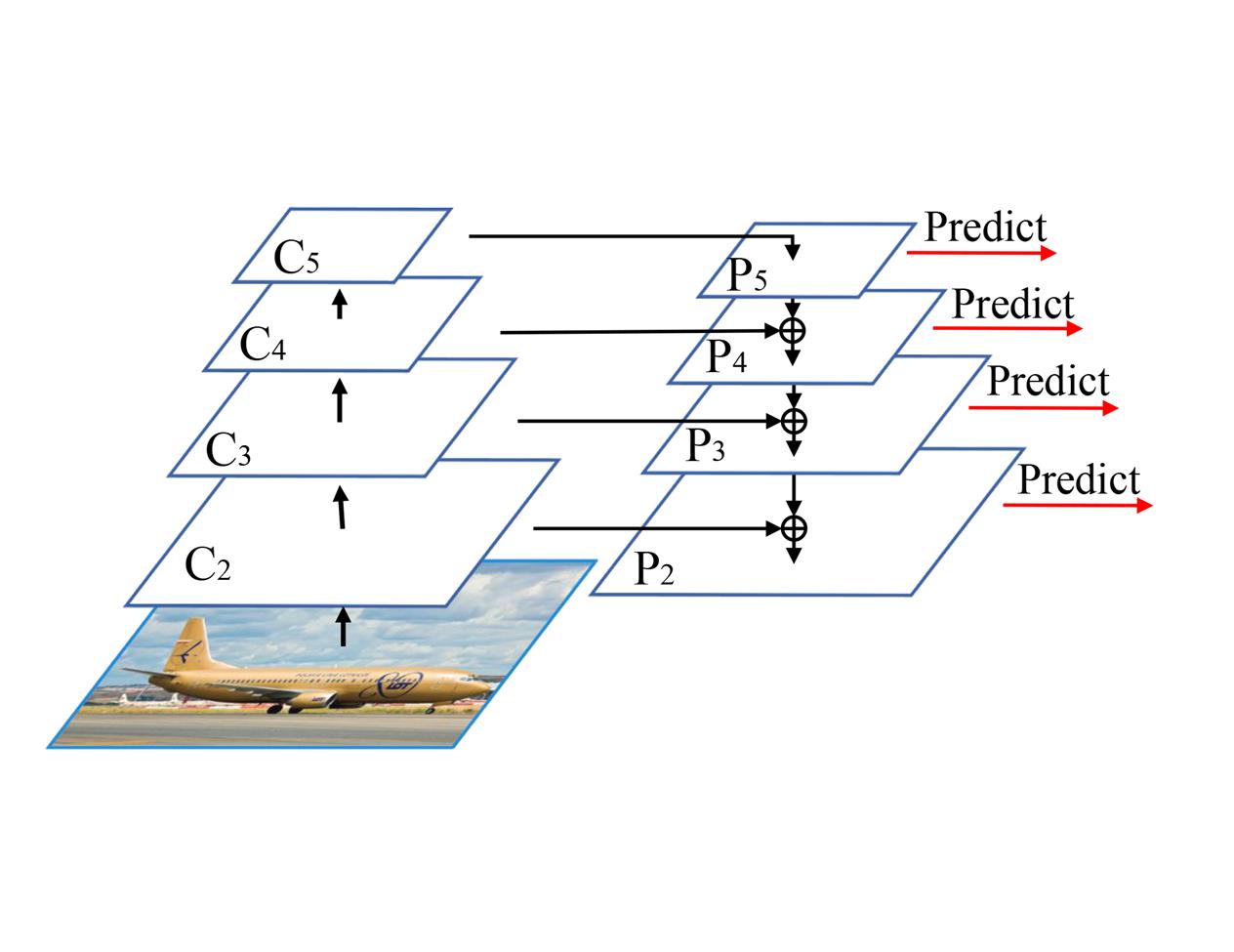}}
	\subfigure[ ]{
		\label{figFPN:b} 
		\includegraphics[width=0.48\linewidth]{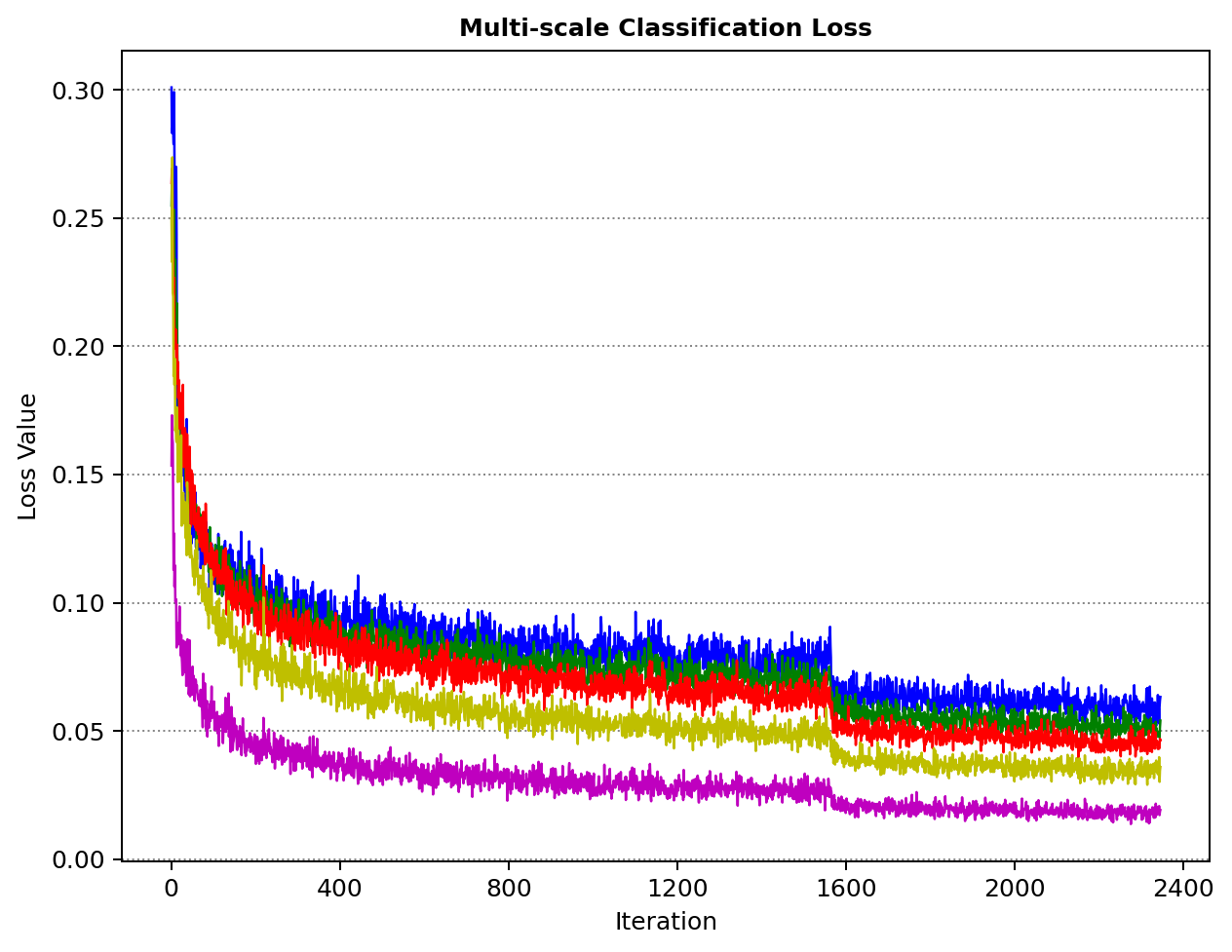}}
	\subfigure[ ]{
		\label{figFPN:c} 
		\includegraphics[width=0.48\linewidth]{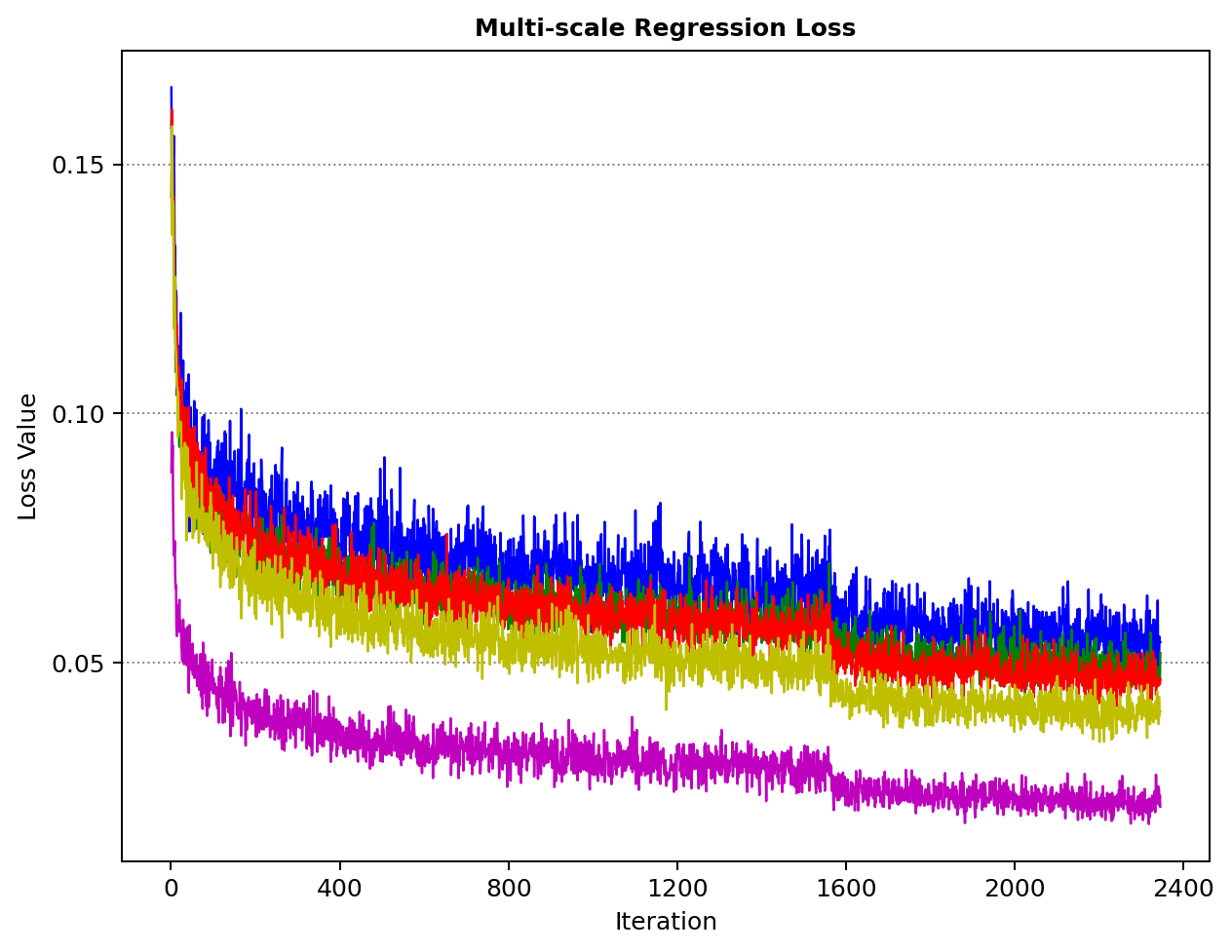}}
	\subfigure[ ]{
		\label{figFPN:d} 
		\includegraphics[width=0.48\linewidth]{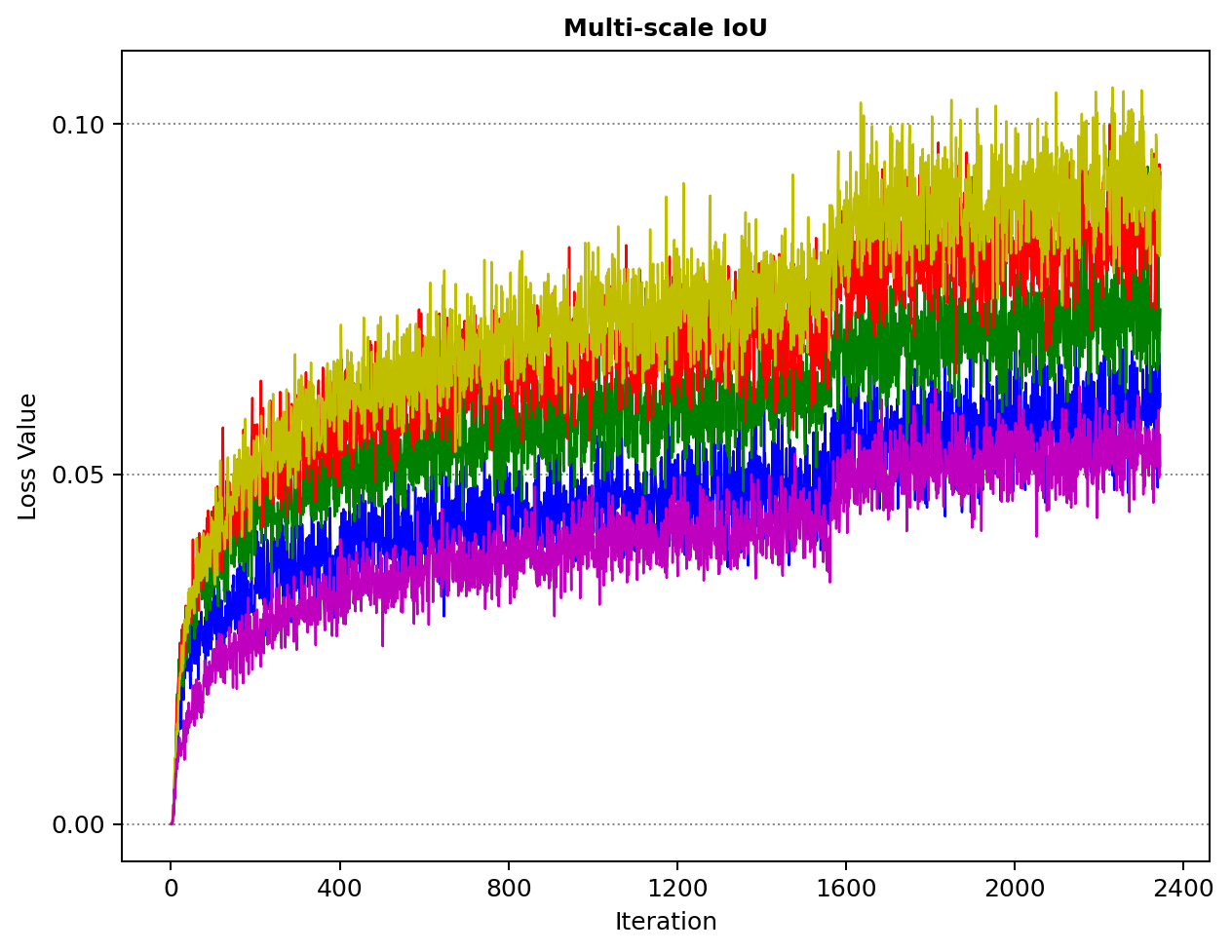}}
	\caption{An illustration of the imbalance problem in multi-scale detectors training.
		(a): In FPN~\cite{lin2017feature}, prediction is performed independently from the pyramid features at different levels.
		(b), (c), (d): The training losses of classification, regression, and IoU through 5 scale predictions, where IoU only counts the value that is not involved in backpropagation.
		The \textbf{blue}, \textbf{green}, \textbf{red}, \textbf{yellow}, and \textbf{pink}
		curves represent the prediction loss every 100 iterations of 5 feature scales from large to small.	
		The losses are trained through RetinaNet~\cite{lin2017focal} ResNet-50~\cite{he2016deep}, 1x schedule on MMDetection~\cite{mmdetection}.
	}
	\label{figFPN}
\end{figure}

As illustrated in Fig~\ref{figFPN:a}, multi-scale feature maps are created by propagating the high-level semantical information into low-level in FPN.
Then predictions (recognition and localization) are independently made on each level while
the loss of all scales is summated without distinction for backpropagation~\cite{lin2017feature}.
Similar to the multi-task training of classification and regression in object detection~\cite{CaoCLL20},
we observe that the loss value is uneven and unstable between each scale level in Fig~\ref{figFPN:b}~\ref{figFPN:c}~\ref{figFPN:d},
which can summarize into three points:
\textbf{(1)} Although the total loss of each scale is averaged based on the number of samples, the larger-scale feature map level has a greater loss.
\textbf{(2)} The fluctuation of the regression loss value is greater than that of the classification branch.
\textbf{(3)} The value of IoU value~\cite{IoU16} (not involved in backpropagation) fluctuates more violently, and it performs better at the mid-scale level.

Based on the illuminating observations, we believe that \textbf{training multi-scale detectors is also a type of multi-task learning},
in addition to classification and regression.
In multi-task learning,
the gradient norm of each task is different, and the task with a large gradient will occupy a dominant position~\cite{Multitask_uncertainty}.
If multi-loss is not balanced, the overall performance may be suboptimal~\cite{Multitask_Prioritization}.
Therefore, we argue that \textbf{training multi-scale detectors also suffers from objective imbalance}.
The common objective imbalance in object detection is training classification and regression tasks simultaneously,
which received relatively little attention~\cite{Imbalance}.
The conventional way to solve it is to weight the tasks~\cite{Multitask_Prioritization}.
Recent approaches~\cite{CaoCLL20,Modulating21} employ the correlation between the classification and regression tasks to assign the weights.
However, training the detectors of each scale is the same task in multi-scale training. It is difficult to determine the weight of each scale level.

In this work, we aim to analyze and alleviate the pointed imbalance problem of multi-scale detectors training.
Inspired by the uncertainty weighting~\cite{Multitask_uncertainty}, we propose Adaptive Variance Weighting (AVW) to balance multi-scale loss.
Unlike previous works~\cite{KLL19,QiLearning20} that estimate the standard deviation through extra networks,
AVW calculates the statistical variance of the training loss in each level to measure their importance.
The loss contributed by the important (high variance reduction rate) scale level is enhanced during training.
For further improvement, we develop a novel Reinforcement Learning Optimization (RLO).
We regard multi-scale detectors as multi-agents, which seek the optimal decision in different training stages.
RLO dynamically decides the optimal scheme for adapting to different phases in model training.
In particular, our methods introduce no extra computational complexity and learnable parameters for backpropagation.

We implement FCOS~\cite{FCOS19}, RetinaNet~\cite{lin2017focal}, and Faster R-CNN~\cite{ren2016faster} with ResNet-50 and ResNet-101~\cite{he2016deep} as the baselines in our experiments to investigate the impacts of our methods.
Experimental results on the MS COCO~\cite{lin2014microsoft} and Pascal VOC~\cite{voc} benchmark show that our approaches can consistently boost the performance over the baseline detectors.
The inference time of the proposed method is the same as the baseline and the extra training burden is negligible.
The main contributions of our work are summarized as follows:

\begin{itemize}
	\item We revisit the recent prominent approaches for the common detection imbalance and carefully analyze the objective imbalance of multi-scale detector training.
	\item We propose a simple yet effective Adaptive Variance Weighting (AVW) to balance multi-scale loss without extra computational burden.
	\item We introduce a Reinforcement Learning Optimization (RLO) to optimize the weighting scheme dynamically. RLO is a novel combination of reinforcement learning and deep detector training.
	\item We evaluate the effectiveness of our proposed methods on MS COCO and Pascal VOC over state-of-the-art FPN-based detectors.
\end{itemize}

\section{Related Work}

\textbf{Multi-scale detection.}
Object detection based on deep learning is usually divided into two-stage detectors~\cite{he2017mask,ren2016faster} and one-stage detectors~\cite{liu2016ssd,lin2017focal}.
More recently, anchor-free detectors~\cite{FCOS19,CenterNet} remove the pre-defined anchor boxes and also achieve excellent performance.
The detectors of almost all frameworks adopt FPN-based multi-scale detection to solve scale imbalance.
FPN~\cite{lin2017feature} merges multi-scale feature maps by propagating the semantical information from high levels into lower levels via a top-down pathway,
while feature-level imbalance appears~\cite{pang2019libra}.
To solve this issue, PANet~\cite{liu2018path} introduces an extra bottom-up pathway for enhancing the low-level information.
Then Libra R-CNN~\cite{pang2019libra} investigates a balanced feature pyramid to pay equal attention to multi-scale features.
After that, AugFPN~\cite{guo2020augfpn} fully extract multi-scale context features to augment the representation abilities of pyramid features.
Cao $\textit{et al.}$~\cite{cao2020attention} design an attention-guided context extraction model to explore large contextual information.
CE-FPN~\cite{CE-FPN} proposes a series of channel enhancement methods that promote performance significantly.
These methods of solving feature-level imbalance commonly stack complex model architectures, which increase the computational burden.
\emph{
	Therefore, we choose the FPN paradigm to study multi-scale detector training.
}

\textbf{Balanced Detector Training.}
Compared with model architectures, alleviating imbalance in the training process is also crucial to object detection~\cite{pang2019libra}.
To balance the foreground-to-background samples, 
OHEM~\cite{OHEM16} selects hard samples according to the confidences automatically.
Considering the efficiency and memory, IoU-based sampling~\cite{pang2019libra} associates the hardness of the examples with their IoUs.
Focal Loss~\cite{lin2017focal} dynamically assigns more weight to the hard examples.
GHM \cite{GHM19} penalizes the loss of a sample if there are many samples with similar gradients.
As for the objective imbalance in detection, the most common solution is weighting the tasks about classification and regression~\cite{Imbalance}.
Several recent studies have improved the traditional methods of manually weighting classification and regression tasks.
CARL~\cite{CaoCLL20} extracts the correlation between the classification and regression tasks to promote regression loss.
Inspired by multi-task learning methods~\cite{Multitask_Prioritization,Multitask_uncertainty},
KL-Loss~\cite{KLL19} attenuates the regression loss of uncertain samples through a trainable weight.
In follow-up research, Qi $\textit{et al.}$~\cite{QiLearning20} adopt a multiple layer perception to train the weights for all the samples and tasks.
The multi-task learning methods for object detection introduce additional layers in the training phase.
\emph{
	Different from the above approaches, we aim to alleviate the objective imbalance of multi-scale detector training.
}

\textbf{Reinforcement learning for object detection.}
There are few researches on the combination of reinforcement learning and object detection.
Active and sequential search methods~\cite{RL15,RL16,RL17,rlrpn18} provide a new approach to perform object detection through reinforcement learning,
which aims to alleviate the object location imbalance~\cite{Imbalance}.
Mathe$\textit{et al.}$~\cite{RL16} use policy gradients to learn RoIs and stop searching automatically.
In~\cite{RL15}, a class-specific agent searches the entire image and performs a sequence of bounding box transformations to localize the object.
And Kong$\textit{et al.}$~\cite{RL17} propose a joint multi-agent Q-learning system to improve the localization results.
To circumvent the issue of training multiple search policies,
RLRPN~\cite{rlrpn18} introduces a sequential region proposal network that refines the search policy with the detector jointly.
Later, PAT~\cite{LiuPay20} locates a sub-region that probably contains relevant objects through an intelligent agent.
These methods have transformed or increased the model architectures of object detection.
\emph{
	Our reinforcement learning strategy has not changed the network architecture of the basic detector, which can be easily embedded into existing detectors.
}

\section{Proposed Methods}

\subsection{Problem Formulation}
\label{sec:ana}

\begin{figure*}[width=2\linewidth]
	\centering
	\includegraphics[width=0.9\linewidth]{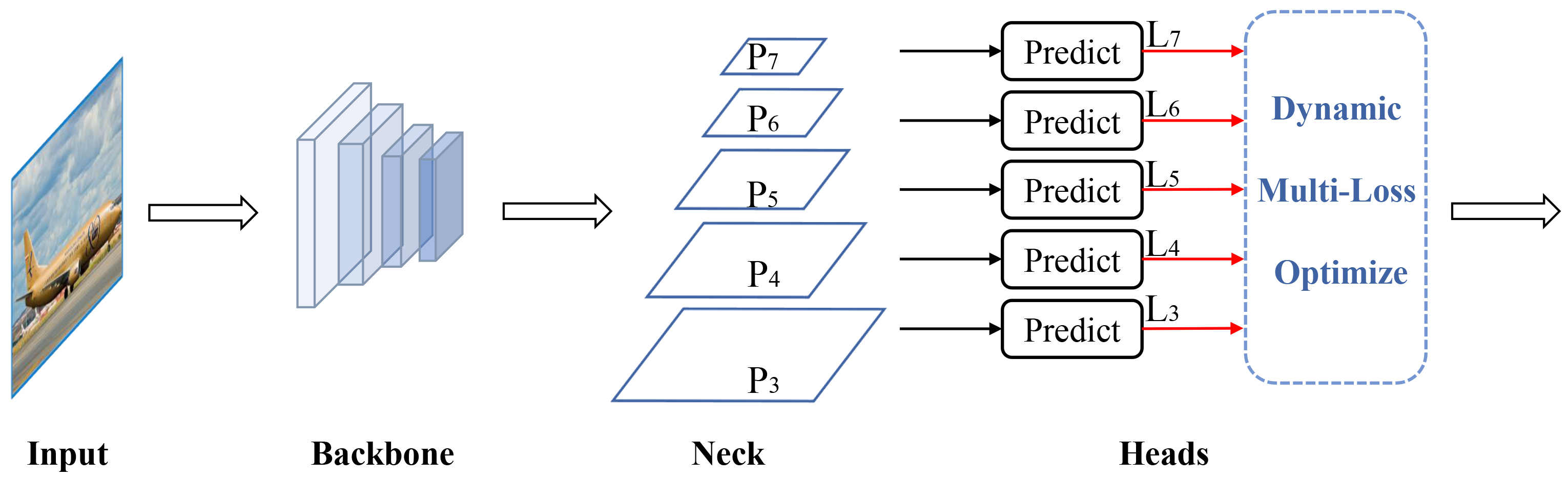}
	\caption{
		An overview of our learning strategy.
		Following MMDetection~\cite{mmdetection}, we omit RPN for simplicity, and the detector consists of three main components: Backbone, Neck and Head.	
		The Head contains DenseHead and RoIHead so that the pipeline is suitable for both one-stage detector and two-stage detector.
		The multi-scale loss produced by Head is input into our algorithm for optimizing dynamically.
		Then the adjusted loss of each scale is summed for backpropagation according to the default configuration of the framework.
		Our proposed Adaptive Variance Weighting (AVW) and Reinforcement Learning Optimization (RLO) are two algorithm configurations for multi-loss optimization.
	}
	\label{figmethod}
\end{figure*}

The overall pipeline is shown in Fig.~\ref{figmethod}, we adopt the description of architecture in MMDetection~\cite{mmdetection}.
Following the framework, we use the FPN as the default configuration for multi-scale detection.
In FPN, feature maps from the backbone output are unified into 256 channels by 1x1 convolution in lateral connection.
We take the RetinaNet of 5 levels pyramid as an example to illustrate our methods.
So that pyramid feature produced by FPN is denoted as $P_i \in \left\{P_3,P_4,P_5,P_6,P_7\right\}$,
and the corresponding loss is $ L_i \in \left\{L_3,L_4,L_5,L_6,L_7\right\}$.
Our study focuses on balancing multi-scale loss, so the classification loss $L_{cls,i}$ and regression loss $L_{reg,i}$ are all regarded as $L_i$.

We first attempt to analyze the causes of multi-scale training imbalance.
The main difference between the detectors of each scale lies in the number and size of samples.
One-stage detectors consider all positions on the image as potential objects and generate samples in each grid cell,
such as RetinaNet~\cite{lin2017focal}, FCOS~\cite{FCOS19}.
In the feature pyramid, the size of the feature map is halved in turn.
Therefore, the number of samples from $P_3$ to $P_7$ decreases exponentially.
Besides, the hierarchical features capture different scale information.
Specifically, shallow spatial-rich features have high resolution and small receptive fields that are suitable for small objects,
while low-resolution features for large objects.
It is the scale difference that causes the possible range inconsistencies among different losses.
As for two-stage detectors, which adopt Region Proposal Networks (RPN) to generate the interested region proposals and assign them to the detection head of each scale.
We observe that the classification loss of each scale is more balanced than that of the one-stage detectors.
But the fixed size assignment~\cite{lin2017feature} still leads to the difference in the regression loss value.

To solve this objective imbalance, we adjust the multi-scale loss by weighting factors, which can be mathematically defined as
\begin{equation}
L_{total} = \sum_{i=3}^{7}w_i L_{i},
\label{eqwL}
\end{equation}
where $L_{total}$ is the total loss for backpropagation. 
The most common approach in multi-task learning is to linearly combine the losses by manual weighting,
which may not address the objective imbalance of the tasks~\cite{Imbalance}.
It is significant how to determine the appropriate loss weight for the current training phase.
Hence, we propose two methods to dynamically optimize $w_i$, which will be described in detail following.

\subsection{Adaptive Variance Weighting}
\label{sec:AVW}

There exist two opposite views on loss and gradient.
Some studies~\cite{OHEM16,lin2017focal} prefer a more significant magnitude of loss and gradient to accelerate training,
while some others~\cite{KLL19,CaoCLL20} attach importance to the samples that have performed better.
To avoid this disagreement, we introduce the statistical variance of the loss to measure the importance of each scale inspired by~\cite{Multitask_uncertainty}.

The multi-task learning method~\cite{Multitask_uncertainty} demonstrates two types of uncertainty: Data-dependent and Task-dependent.
In object detection, the loss value depends on the qualities of the input samples~\cite{CaoCLL20}.
So we introduce a relatively large iteration interval to exclude the influence of the Data-dependent uncertainty.
Different from the 'variance' trained through fully connected layers~\cite{QiLearning20},
we calculate the statistical variance of each scale level in the iteration interval.
Under different loss values, the decrease in variance means that the training results are more stable and confident.
We believe that the scale levels with high statistical variance reduction rate are more important,
whose loss needs to be enhanced during training to prevent other larger gradients from occupying a dominant position.
We treat iteration as a time dimension and redefine the interval loss of each scale level as follows:
\begin{equation}
\mathcal{L}_{i,t} = \sum_{m=(t-1)\times \alpha + 1}^{t\times \alpha}L_{i,m}, \quad  t=1,2,3,...\quad \mathrm{s.t.} \quad m \in (0, T],
\label{eqsumL}
\end{equation}
where $m$ is the current iteration, $T$ denotes total iterations,
$i\in \left\{3,4,5,6,7\right\}$ indicates the index of the scale level,
$t$ denotes the time sequence,
and $\alpha$ is the interval step.
The impacts of $\alpha$ are analyzed in Sec~\ref{sec:abla}.
Then the variance of each loss value per iteration interval can be calculated as
\begin{equation}
Var_{i,t} = \frac{ 1}{\alpha - 1}\sum_{m=(t-1)\times \alpha + 1}^{t\times \alpha} (L_{i,m} - \frac{\mathcal{L}_{i,t}}{n})^2,
\label{eqvar}
\end{equation}
\begin{equation}
r_{i,t} = \frac{Var_{i,t-1} - Var_{i,t}} {Var_{i,t-1} + \epsilon} ,
\label{eqrate}
\end{equation}
where $r_{i,t}$ denotes the rate of decrease in variance,
and $\epsilon=0.00001$ is a small value to avoid numerical instability.
The initial value of $r_{i,t}$ and $Var_{i,t-1}$ are set to 0, 1 respectively.

We believe that the two levels with the maximal $r_{i,t}$ are more important than the others,
whose weights are selected to be increased.
The weight of each scale level $w_i$ is updated at the end of each iteration interval.
$w_i$ is initialized as 1.
We have studied the number of levels that require enhancement through ablation experiments.
Intuitively, 2 is an appropriate number, and the experimental results prove that it is indeed optimal.
Besides, increasing the loss of all scale levels is similar to enlarging the learning rate,
which is not in the scope of our research.
$w_i$ of these two scale levels are adjusted to a larger value as
\begin{equation}
w_j =\left\{\begin{array}{ll}{1 + \lambda\frac{\mathcal{L}_{j,t}}{\sum\mathcal{L}_{i,t}}} & {\text {if j is the selected level,} } \\ {1} & {\text { otherwise, }}\end{array}\right.
\label{eqweight}
\end{equation}
where $j$ denotes the index of the selected two scale levels, and $\lambda$ is an amplification factor.
Empirically, $\lambda$ is set to 1.5 for the level with the maximal $r_{t}$ and 1 for the other.
And $w_i$ of the remaining level is adjusted to 1.
We name this method Adaptive Variance Weighting (AVW).

\subsection{Reinforcement Learning Optimization}

In terms of reinforcement learning, exploration and exploitation are both significant to the system~\cite{RLbook}.
The agent has to exploit what it already knows to obtain reward and explore better action selections in the future. 
In trying a series of actions, the agent progressively favors those that appear to be best.
AVW is essentially a subjective greedy strategy,
which may not be suitable for the entire training epochs.
Therefore, we attempt to introduce the exploration to obtain more long-term benefits.

The whole detector can be seen as a reinforcement learning system.
We regard multi-scale detectors as multi-agents, which seek the optimal decision in different training stages.
The agents sense the state of the environment continuously and perform actions,
which will return a real-valued reward signal to agents for awarding or punishing the selection.
The policy guides multi-agents to execute the next decision.
We will elaborate on the elements in the reinforcement learning system as follows:

\textbf{State}.
The states are a set of training results during an iteration interval.
According to the calculation results of Eq~\ref{eqsumL},~\ref{eqvar},
We define the current state as
\begin{equation}
\mathcal{S}_t = \left\{ \mathcal{L}_{i,t} ,\quad \mathcal{L}_{i,t-1},\quad Var_{i,t}, \quad Var_{i,t-1} \right\}.
\label{eqstate}
\end{equation}
As shown in Fig~\ref{figFPN:d}, we do not include IoU value in the state due to its severe fluctuation,
which does not contribute much to improving the performance in our task.

\textbf{Action}.
We expect more actions to find the optimal strategy.
AVW is a possible action in a set of actions.
As mentioned in Sec~\ref{sec:AVW}, the two opposite views on loss and gradient both make sense for detection.
Inspired by them, we believe that enhancing levels of small loss is conducive to performance improvement in some phases of network training.
And it is better to enhance the level with a large loss in some other cases.
We can implement these two concepts similar to the procedure of AVW.
Besides, the learning rate is reduced a lot in the later phase of model training.
In the case of near convergence, excessive stimulation of weights will lead to the degradation of detection performance.
It may be better not to increase the value of the weight.

In summary, there are four proposed schemes to update $w_i$.
We denote the action set as $\mathcal{A} = \left\{a_0,a_1,a_2,a_3\right\}$.
Multi-agents decide an action $A_t$ at time $t$ to execute through the policy.
And each action in $\mathcal{A}$ is defined as follows:

$a_0: $ Select the two levels with the maximal $r_{i,t}$, and update $w_i$ by Eq~\ref{eqweight}.

$a_1: $ Select the two levels with the minimal $\mathcal{L}_{i,t}$, and update $w_i$ by Eq~\ref{eqweight}.

$a_2: $ Select the two levels with the maximal $\mathcal{L}_{i,t}$, and update $w_i$ by Eq~\ref{eqweight}.

$a_3: $ Each $w_i$ is updated to 1.

\textbf{Reward}.
The reward function is critical to agent building, which defines what are the good and bad events for the agent~\cite{RLbook}.
In the reinforcement learning system, multi-agents should improve overall performance rather than individual optimization.
The intuitive result is that the total loss has decreased in the training process.
To this end, we design the reward function as
\begin{equation}
R_{A_{t-1}}\left(\mathcal{S}_{t-1}, \mathcal{S}_t \right) =
\operatorname{sign}\left( \sum\mathcal{L}_{i,t-1} - \sum\mathcal{L}_{i,t} \right),
\label{eqreward}
\end{equation}
where $\mathcal{S}_t$ denotes the state at time $t$, $\mathcal{S}_{t-1}$ is the previous one.
$R_{A_{t-1}}$ is obtained at $t$, which represents the reward signal of the action $A_{t-1}$.
The initial value of the reward is set to 0.
A positive reward is returned if the total loss between two adjacent states drops.
Otherwise, its value is less than or equal to zero.
For the same reason, we do not consider IoU rewards.

\textbf{Policy}.
We devise a probabilistic decision policy to guide multi-agents.
Each action in $\mathcal{A}$ is assigned a selection probability.
At time $t$, the probability of performing action $a_k$ is $p_k$, which is defined as
\begin{equation}
P_{\mathcal{A}}\left(A_{t}=a_k  \right)= p_{k,t},  \quad  k=0,1,2,3,
\label{eqP}
\end{equation}
where $\mathcal{P}_\mathcal{A}$ denotes the set of probabilities $\left\{p_0,p_1,p_2,p_3\right\}$.
To make a trade-off between exploration and development, we do not introduce any prior knowledge.
The initial values of $p_{k}$ are all $0.25$.

According to the reward signal, $\mathcal{P}_\mathcal{A}$ is continuously updated.
With the progress of backpropagation, the overall loss will gradually decrease.
We believe that the action damages the model training if it returns a non-positive reward value.
So we punish this kind of action by reducing its selection probability.
And the probability of the action that produces a positive reward will be increased.
The update process of $\mathcal{P}_\mathcal{A}$ at time $t$ is defined as
\begin{equation}
\begin{aligned}
&p_{k,t} = \operatorname{brelu}\left(  p_{k,t-1} \pm \gamma \right) , \quad \text {if } A_{t-1}=a_k,  \\
&p_{q,t} =  p_{q,t-1} \frac{1-p_{k,t}}{1-p_{k,t-1}}, \quad  p_{q,t} \in \mathcal{P}_\mathcal{A} - p_{k,t}, \\
&\operatorname{brelu}(x) = min(max(x,\beta_{min}),\beta_{max}).
\end{aligned}
\label{equpdateP}
\end{equation}

In Eq~\ref{equpdateP}, $p_{k,t}$ is the probability of the previous action $A_{t-1}$, which requires award or punishment first through a small factor $\gamma$.
$\pm$ indicates that if $R_{A_{t-1}} >0$ performing $+$, otherwise $-$.
After that, the remaining three probabilities $p_{q,t}$ are updated proportionally, which asserts the sum of all probabilities is 1.
And $\beta_{min}, \beta_{max}$ are the boundary value of $p_{k,t}$ if outliers.

We name this method Reinforcement Learning Optimization (RLO).
In Algorithm~\ref{algorithmW}, we have summarized the procedure for generating weights of multi-scale loss.
RLO can be easily embedded in the multi-scale detector training process. 
AVW is a specific case of RLO, which constantly performs action 0 with a 100\% probability.

\begin{algorithm}[t]
	\caption{Reinforcement Learning Optimization} 
	{\bf Input:}\\ 
	\hspace*{0.2in}
	$\mathcal{S}$: The state set calculated by the training losses.\\
	\hspace*{0.2in}
	$\mathcal{A}$: The action set for generating weights of multi-scale loss.\\
	\hspace*{0.2in}
	$\alpha$: Update interval factor.\\
	{\bf Output:}\\ 
	\hspace*{0.2in}
	Trained object detector $\mathcal{D}$.
	\begin{algorithmic}[1]
		\State {\bf Begin}
		\State Initialize counter $t$, reward signal $R$, multi-scale weights $w_i$, probabilities set $\mathcal{P}_\mathcal{A}$
		\For{$m=0$ to max\_iter}
		\State Train object detector $\mathcal{D}$ by Eq~\ref{eqwL}, and obtain the output multi-loss for $\mathcal{S}$
		\If{$m\% \alpha ==0$}    \quad   \#Execute every $\alpha$ iteration
		\State $t+=1$
		\State Calculate current state $\mathcal{S}_t$ by Eq~\ref{eqsumL},~\ref{eqvar}
		\State Calculate reward $R$ by Eq~\ref{eqreward}
		\State Update $\mathcal{P}_\mathcal{A}$ according to $R$ by Eq~\ref{equpdateP}
		\State Select $A_t =a_k$ according to the probability $\mathcal{P}_\mathcal{A}$
		\State Execute $A_t$ to update $w_i$ in Eq~\ref{eqwL}
		\EndIf
		\EndFor
		\State {\bf End}
	\end{algorithmic}
	\label{algorithmW}
\end{algorithm}

\section{Experiments}

\subsection{Dataset and Evaluation Metrics}

We perform our experiments on MS COCO~\cite{lin2014microsoft} and Pascal VOC~\cite{voc} benchmark.
It contains 80 categories and consists of 115k images for training (train-2017), 5k images for validation (val-2017) and 20k images in test-dev.
We train all the models on train-2017 and report results on val-2017.
The performance metrics follow the standard COCO-style mean Average Precision (mAP) metrics
under different Intersection over Union (IoU) thresholds, ranging from 0.5 to 0.95 with an interval of 0.05.
And Pascal VOC~\cite{voc} covers 20 common categories in everyday life.
We merge the VOC2007 trainval and VOC2012 trainval split for training and evaluate on VOC2007 test split.
The mean Average Precision (mAP) for VOC is under 0.5 IoU.

\subsection{Implementation Details}

For fair comparisons, all experiments are implemented based on MMDetection v2~\cite{mmdetection}.
MMDetection has been upgraded to v2 with higher baseline performance than v1.
We train the detectors with the resolution of (1333, 800) on 4 NVIDIA Quadro P5000 GPUs (2 images per GPU) for 12 epochs (1x schedule).
Our method is a training strategy, and we will verify their effectiveness and generalization on different detectors and backbones.
So FCOS~\cite{FCOS19}, Faster R-CNN~\cite{ren2016faster}, and RetinaNet~\cite{lin2017focal} are chosen as the baseline detectors,
which represent anchor-free, two-stage, and one-stage detectors respectively.
And the learning rate of them are set to 0.01, 0.01, and 0.005 respectively.
The learning rate is dropped by 0.1 after 8 and 11 epochs.
As for Pascal VOC, the total epochs are 4 and the learning rate is dropped by 0.1 after 3 epochs.
Two classical networks ResNet-50 and ResNet-101~\cite{he2016deep} are adopted as backbones.
And the multi-scale module (Neck) is FPN~\cite{lin2017feature}.
The other settings follow the basic framework of MMDetection as default.

In hyperparameter settings,
we set $\lambda$ in Eq~\ref{eqweight} to 1.5 and 1 for the two selected levels.
The larger value of $\lambda$ will decrease model performance.
In Eq~\ref{equpdateP}, the award/punishment factor $\gamma$ is set as 0.01,
because we expect to explore some training knowledge from gradual probability changes.
To prevent outliers in probability values, we set $\beta_{min}=0.1$ and $\beta_{max}=0.9$.
The iteration interval $\alpha=100$ and the number of selected levels $N$ is 2, which are analyzed in our ablation experiments.

\subsection{Main Results}

\begin{table*}[width=2\linewidth]
	\centering
	\caption{Comparison with baselines and state-of-the-art methods. The symbol '*' means the re-implemented results on mmdetection.}
	\label{tabresults}
		\begin{tabular}{l|c|c|ccccc}
			\toprule
			Method         &     Backbone    & \quad $AP$ \quad \quad  & $AP_{50}$ & $AP_{75}$ & $AP_{S}$ & $AP_{M}$ & $AP_{L}$ \\
			\midrule
			\textbf{\textit{State-of-the-art:}}	\\
			Guided Loss~\cite{Guided}          &    ResNet-50     & 37.0        & 56.5 & 39.2 & 20.3 & 40.5 & 49.5 \\
			FSAF~\cite{FSAF}                   &    ResNet-50     & 37.2        & 57.2 & 39.4 & 21.0 & 41.2 & 49.7 \\
			Foveabox*~\cite{Foveabox}          &    ResNet-50     & 36.5        & 56.0 & 38.6 & 20.5 & 39.9 & 47.7 \\ 
			Faster RCNN+Balanced-L1*~\cite{pang2019libra}&ResNet-50& 37.4       & 58.2 & 41.0 & 21.5 & 41.0 & 48.3 \\
			Faster RCNN+PAFPN*~\cite{liu2018path}&    ResNet-50   & 37.5        & 58.6 & 40.8 & 21.5 & 41.0 & 48.6 \\
			RetinaNet + GHM*~\cite{GHM19}      &    ResNet-50     & 36.9        & 55.5 & 39.1 & 20.3 & 40.4 & 48.9 \\
			RetinaNet + GHM*~\cite{GHM19}      &    ResNet-101    & 39.0        & 58.0 & 41.6 & 22.1 & 42.8 & 51.9 \\
			\midrule
			\midrule
			\textbf{\textit{Baseline} + \textit{Ours:}}	\\
			FCOS*~\cite{FCOS19}                &    ResNet-50     & 36.6        & 56.0 & 38.8 & 21.0 & 40.6 & 47.0 \\
			FCOS w/ \textbf{RLO}               &    ResNet-50     & 37.0        & 56.3 & 39.1 & 21.0 & 40.8 & 47.5 \\
			\midrule
			FCOS*~\cite{FCOS19}                &    ResNet-101    & 39.0        & 58.2 & 42.1 & 22.7 & 43.3 & 50.3 \\
			FCOS w/ \textbf{RLO}               &    ResNet-101    & 39.3        & 58.9 & 41.9 & 22.6 & 42.9 & 51.0 \\
			\midrule
			Faster RCNN*~\cite{ren2016faster}  &    ResNet-50     & 37.2        & 57.9 & 40.6 & 21.3 & 40.8 & 47.9 \\
			Faster RCNN w/ \textbf{RLO}        &    ResNet-50     & 37.7        & 58.8 & 40.7 & 22.1 & 41.1 & 48.2 \\
			\midrule
			Faster RCNN*~\cite{ren2016faster}  &    ResNet-101    & 39.2        & 59.6 & 42.9 & 22.5 & 43.3 & 51.4 \\
			Faster RCNN w/ \textbf{RLO}        &    ResNet-101    & 39.5        & 60.2 & 43.0 & 22.3 & 43.9 & 52.0 \\
			\midrule
			RetinaNet*~\cite{lin2017focal}     &    ResNet-50     & 36.1        & 55.4 & 38.5 & 19.8 & 39.7 & 47.1 \\
			RetinaNet w/ \textbf{RLO}          &    ResNet-50     &\textbf{36.9}& 56.3 & 39.1 & 20.8 & 40.1 & 48.5 \\
			\midrule
			RetinaNet*~\cite{lin2017focal}     &    ResNet-101    & 38.2        & 57.4 & 40.7 & 22.2 & 42.5 & 49.9 \\
			RetinaNet w/ \textbf{RLO}          &    ResNet-101    &\textbf{38.9}& 58.0 & 41.5 & 21.6 & 43.1 & 51.2 \\
			\bottomrule
		\end{tabular}
\end{table*}

As shown in Table~\ref{tabresults}, we evaluate our approach with the baselines and state-of-the-art methods.
For fair comparisons, we report the re-implemented results of them on mmdetection and COCO val-2017.
The selected state-of-the-art aim to alleviate the imbalance issue from different training perspectives.
Experimental results show that our approach achieves competitive performance compared to them.
RLO with RetinaNet using ResNet-50 and ResNet-101 as backbone achieves \textbf{36.9} and \textbf{38.9} AP,
which is \textbf{0.8} and \textbf{0.7} points higher than baselines respectively.
And RLO improves Faster R-CNN by 0.5 and 0.3 AP.
As for FCOS, the performance is boosted to 37.0 and 39.3 AP.
We observe that results of $AP_{L}$ are consistently improved,
which indicates that the training of low-resolution feature maps for large objects has improved.

As mentioned in Sec~\ref{sec:ana}, the objective imbalance of multi-scale training in one-stage detectors is more common than the two-stage.
RPN makes the multi-scale loss of two-stage detectors more balanced.
And FCOS adopts IoU-Loss as the regression loss function with fluctuating variance, which impedes our method.
The results indicate that our approach improves the one-stage detectors significantly, which proves our analysis.
And ResNet-101 has a stronger feature representation ability than ResNet-50,
so the performance improvement of ResNet-101 is less than ResNet-50.

Table~\ref{tabvoc} shows the evaluations on Pascal VOC 2007 test.
Experimental results demonstrate the effectiveness of our approaches on performance improvements.
We can believe our method can consistently boost the performance of region-based object detectors.

\begin{table}[!t]
	\centering
	\caption{Comparison with baselines on VOC2007 test.}
	\label{tabvoc}
	\begin{tabular}{l|c|cccccc}
		\toprule
		Methods    & Backbone & $mAP$   \\
		\midrule
		RetinaNet  & ResNet-50 & 77.3         \\
		RetinaNet w/ RLO  & ResNet-50 & \textbf{78.8}\\
		\midrule
		Faster R-CNN  & ResNet-50 & 79.5         \\
		Faster R-CNN w/ RLO  & ResNet-50 & \textbf{80.9}\\
		
		\bottomrule
	\end{tabular}
\end{table}

\begin{figure*}
	\centering
	\begin{minipage}[c]{\linewidth}
		\includegraphics[width=0.3\linewidth]{000000042528_gt.jpg}
		\includegraphics[width=0.3\linewidth]{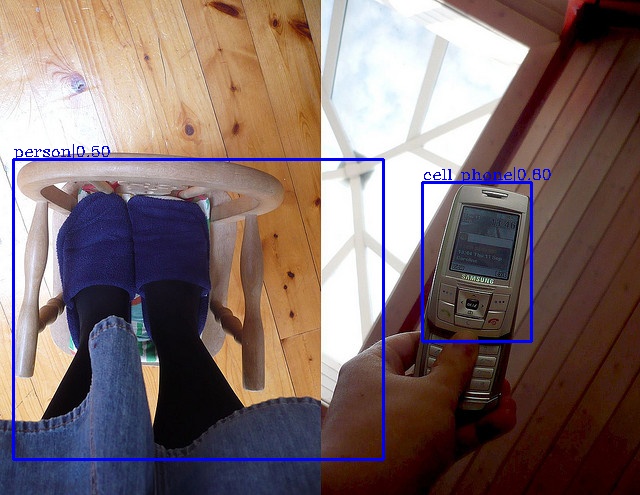}
		\includegraphics[width=0.3\linewidth]{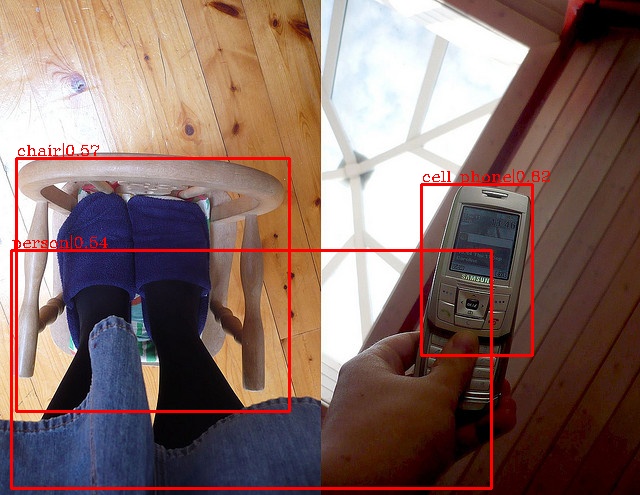}
	\end{minipage}
	\begin{minipage}[c]{\linewidth}
		\includegraphics[width=0.3\linewidth]{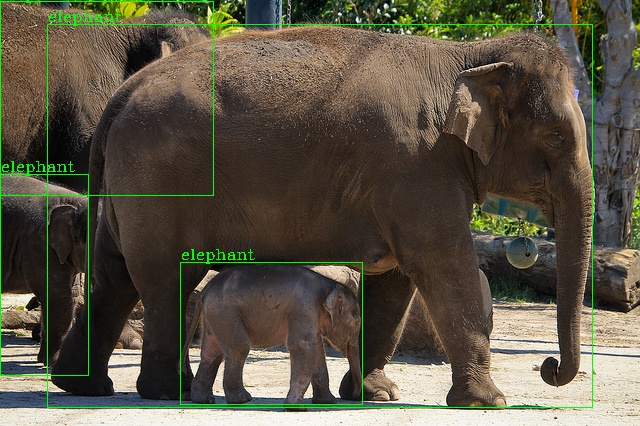}
		\includegraphics[width=0.3\linewidth]{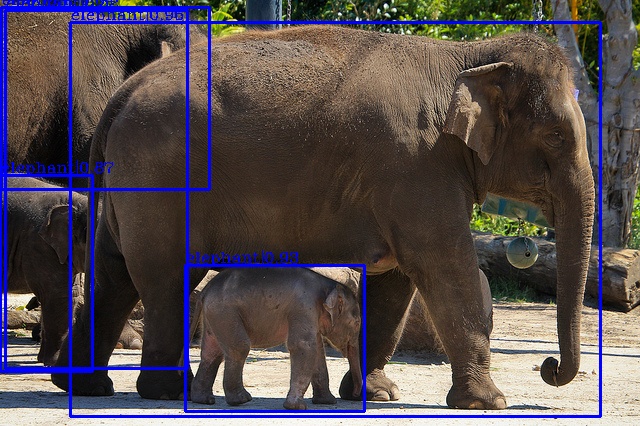}
		\includegraphics[width=0.3\linewidth]{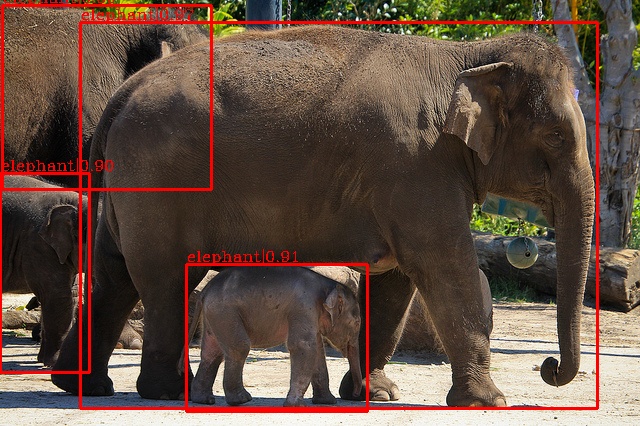}
	\end{minipage}
	\begin{minipage}[c]{\linewidth}
		\includegraphics[width=0.3\linewidth]{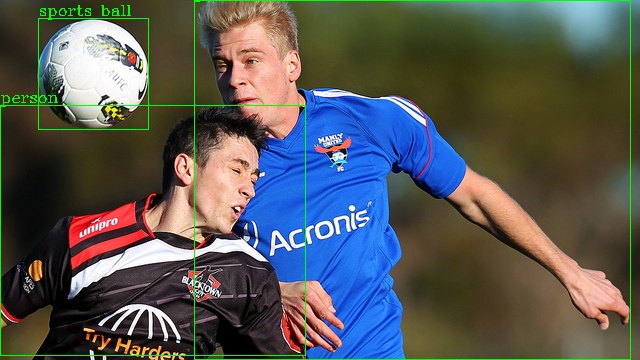}
		\includegraphics[width=0.3\linewidth]{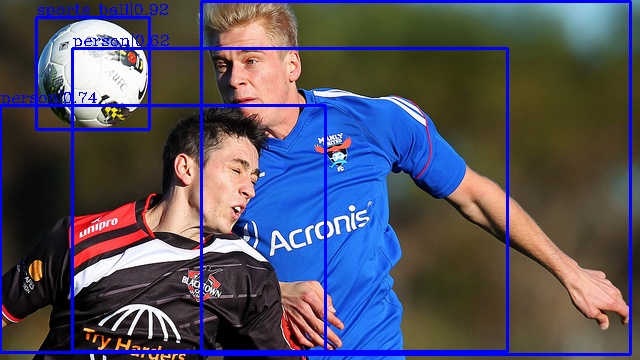}
		\includegraphics[width=0.3\linewidth]{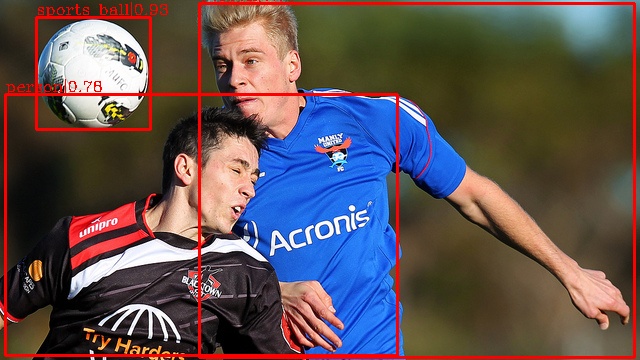}
	\end{minipage}
	\begin{minipage}[c]{\linewidth}
		\includegraphics[width=0.3\linewidth]{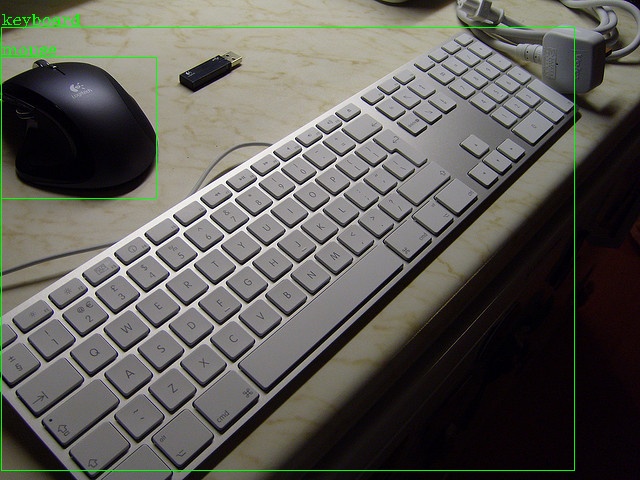}
		\includegraphics[width=0.3\linewidth]{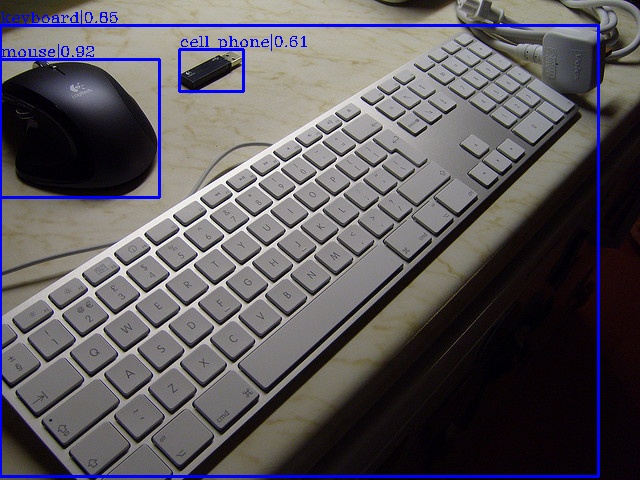}
		\includegraphics[width=0.3\linewidth]{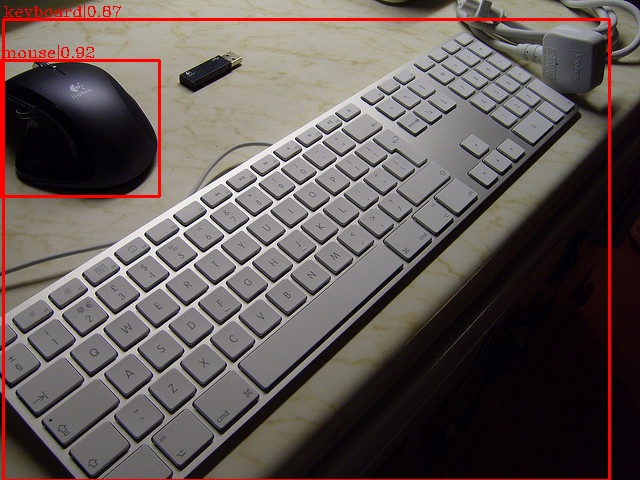}
	\end{minipage}
	\caption{Qualitative results comparison.
		The first column is the ground truth with green bounding boxes.
		The results of the original RetinaNet are listed by the blue bounding boxes, while those of RLO are the red bounding boxes.}
	\label{figimgs}
\end{figure*}

We also show the comparisons of the qualitative results in Fig.~\ref{figimgs}.
Both models are built upon ResNet-50 + FPN and $1\times$ schedule.
The images are chosen from COCO val-2017.
We compare the detection performance with threshold = 0.5.
It can be seen that RetinaNet+RLO generates satisfactory results especially in terms of large objects,
while the original RetinaNet generates inferior results.

\subsection{Ablation Experiments}
\label{sec:abla}

\begin{table*}[width=2\linewidth]
	\centering
	\caption{Effect of each actions on COCO val-2017.}
	\label{tababla}
		\begin{tabular}{cccc|c|ccccc}
			\toprule
			\begin{tabular}[c]{@{}c@{}}action 0 \\ (\textbf{AVW})\end{tabular}  &
			action 1 &
			action 2    &
			\begin{tabular}[c]{@{}c@{}}action 3 \\ (baseline)\end{tabular}  &
			\quad $AP$ \quad \quad   & $AP_{50}$ & $AP_{75}$ & $AP_{S}$ & $AP_{M}$ & $AP_{L}$ \\
			\midrule
			&  &  & \checkmark & 36.1        & 55.4 & 38.5 & 19.8 & 39.7 & 47.1 \\
			\midrule
			\checkmark &  &  & & 36.6        & 55.2 & 39.0 & 19.5 & 40.3 & 49.0 \\
			&  \checkmark & &  & 36.3        & 55.2 & 38.9 & 20.3 & 40.1 & 47.3 \\
			&  & \checkmark &  & 36.0        & 54.8 & 38.4 & 20.3 & 40.0 & 46.2 \\
			& \checkmark & \checkmark &  & 36.4        & 55.3 & 38.7 & 20.0 & 40.0 & 47.6 \\
			\checkmark & \checkmark & \checkmark & \checkmark & 36.9        & 56.3 & 39.1 & 20.8 & 40.1 & 48.5 \\
			\bottomrule
		\end{tabular}
\end{table*}

First, we evaluate the effect of each proposed action of RLO with RetinaNet-ResNet50-FPN on COCO val-2017.
The overall ablation studies are reported in Table~\ref{tababla}.
Running action 3 independently represents the baseline, and action 0 is AVW.
When adjusting the multi-scale weights with a fixed scheme,
AVW performs better than others with an improvement of 0.5 AP,
which illustrates the significance of statistical variance for multi-scale training.
Enhancing the scale levels with large losses (action 2) will harm model performance if implemented alone.
And the combination of action 2 and 3 can also bring improvements, which proves the effectiveness of our reinforcement learning optimization.
\emph{
	Therefore, performing a fixed scheme throughout the training process may lead to sub-optimal.
	Optimizing strategies dynamically during different phases is significant to multi-scale detectors training.
}

Then, we analyze the impact of the hyperparameters.
As mentioned in Sec~\ref{sec:AVW}, the iteration interval $\alpha$ is to exclude the influence of the Data-dependent uncertainty.
We assign various values to $\alpha$ for checking its sensitivity to our algorithm.
Table~\ref{tabalpha} illustrates that the results are better when $\alpha \geq 50$ and achieves the best performance at 100.
If the value of $\alpha$ is too small, the state $\mathcal{S}$ will fluctuate drastically and decrease detection performance.
As for the number of levels that require enhancement $N$,
we have enumerated all the values in Table~\ref{tabnum}.
It shows that 2 is the optimal value of $N$ while
increasing the loss across all scales will impede detection performance.


\begin{table}[!t]
	\centering
	\caption{ Performance comparisons by varying $\alpha$.}
	\label{tabalpha}
		\begin{tabular}{c|cccccc}
			\toprule
			\quad $\alpha$ \quad \quad &  5  & 50  &  100         &  200  & 500 \\
			\midrule
			$AP$    \quad  & 35.9&36.4&\textbf{36.9}& 36.7 & 36.5\\
			\bottomrule
		\end{tabular}
\end{table}

\begin{table}[!t]
	\centering
	\caption{ Performance comparisons by varying $N$, the number of levels that require enhancement.}
	\label{tabnum}
		\begin{tabular}{c|cccccc}
			\toprule
			\quad $N$ \quad \quad &  1  & 2  &  3         &  4 & 5  \\
			\midrule
			$AP$    \quad  & 36.5&\textbf{36.9}& 36.3& 36.0  & 35.9\\
			\bottomrule
		\end{tabular}
\end{table}

Besides, we have observed the trend of probabilities of actions during the training process.
The inclination of RLO is changing at different training phases.
We initialize $\mathcal{P}_\mathcal{A} = \left\{0.25,0.25,0.25,0.25\right\}$ without any prior knowledge.
The probability changes randomly at the beginning of training.
Then $p_0$ and $p_1$ slowly rise in epochs 2 to 6 (total 12),
while $p_2$ dominates in epochs 7 and 8.
After dropping the learning rate at epoch 9,
the detector training approximately converged.
Finally, $p_3$ becomes the maximum among them.
This observation confirms our proposed dynamic optimization.
The concept of dynamic training has also been proved significant in~\cite{Dynamic20}.

\subsection{Inference speed}

\begin{figure}
	\centering
	\includegraphics[width=\linewidth]{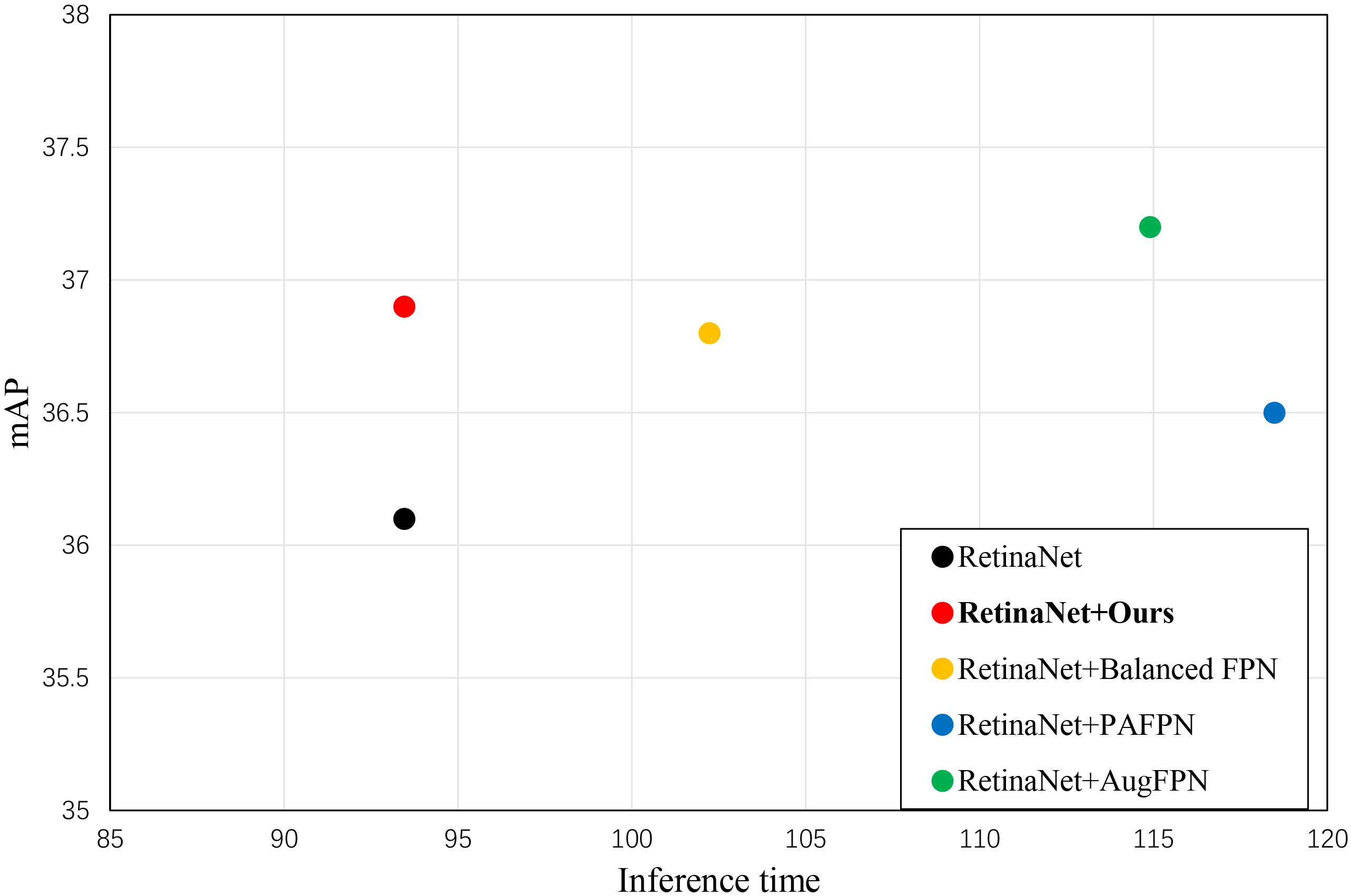}
	\caption{Comparisons of inference time with FPN-based detectors.}
	\label{figfps}
\end{figure}

Finally, we measure the training and testing time of RLO.
When embedding RLO into RetinaNet (ResNet-50 + FPN), the average training time per 50 iterations only increases from \textbf{31.32s} to \textbf{31.46s}.
And the inference time of the proposed method is the same as the baseline.
Fig~\ref{figfps} shows the comparison of inference time with FPN-based methods: Balanced FPN~\cite{pang2019libra}, PAFPN~\cite{liu2018path} and AugFPN~\cite{guo2020augfpn}.
Our approach does not add additional cost during inference,
while other FPN-based approaches generally bring extra computation.
All the runtimes are tested on a single NVIDIA Quadro P5000 with the input of (1333, 800).

\section{Conclusion}

In this paper, we argue that training multi-scale detectors is also a type of multi-task learning in addition to classification and regression,
which suffers from an objective imbalance.
Based on our observation and analysis,
we propose Adaptive Variance Weighting and Reinforcement Learning Optimization to improve multi-scale training.
Extensive experiments show that our approaches can consistently boost the performance over various baseline detectors without extra computational complexity.
We hope that our viewpoint can inspire further researches on multi-scale training and reinforcement learning.

\section*{Declaration of interests}
The authors declare that they have no known competing financial interests or personal relationships that could have appeared to influence the work reported in this paper.

\section*{Acknowledgment}
This work was supported in part by the National Natural Science Foundation of China under Grant 61572214, and Seed Foundation of Huazhong University of Science and Technology (2020kfyXGYJ114).

\bibliographystyle{cas-model2-names}

\bibliography{cas-refs}

\end{document}